\documentclass[a4paper]{article}

\usepackage{INTERSPEECH2018}

\graphicspath{{fig/}}
\newcommand{\imagesep}{\vspace*{-6pt}}
\newcommand{\tablesep}{\vspace*{-6pt}}

\usepackage{booktabs}
\usepackage{tabularx}
\usepackage{multirow}
\newcommand{\mytable}{
    \centering
    \renewcommand{\arraystretch}{1.2}
    }
\newcolumntype{C}{>{\centering\arraybackslash}X}
\newcolumntype{L}{>{\raggedright\arraybackslash}X}
\newcolumntype{R}{>{\raggedleft\arraybackslash}X}
\newcolumntype{P}[1]{>{\raggedright\arraybackslash}p{#1}}

\usepackage{microtype}
\usepackage{url}
\usepackage{xcolor}
\newcommand{\system}[1]{{\footnotesize \textsc{#1}}}

\usepackage{cite}
\title{Visually grounded cross-lingual keyword spotting in speech}
\name{Herman Kamper$^1$ and Michael Roth$^2$}
\address{$^1$E\&E Engineering, Stellenbosch University, South Africa \& $^2$Saarland University, Germany}
\email{kamperh@sun.ac.za, mroth@coli.uni-sb.de}

\usepackage[prependcaption,textsize=scriptsize]{todonotes}
\definecolor{mycolor}{HTML}{FF6600}

\definecolor{mycolor2}{HTML}{66FF00}

\begin{document}

\maketitle
\begin{abstract}
Recent work considered how images paired with speech can be used as supervision for building speech systems when transcriptions are not available. We ask whether visual grounding can be used for \textit{cross-lingual keyword spotting}: given a text keyword in one language, the task is to retrieve spoken utterances containing that keyword in another language. This could enable searching through speech in a low-resource language using text queries in a high-resource language. As a proof-of-concept, we use English speech with German queries: we use a German visual tagger to add keyword labels to each training image, and then train a neural network to map English speech to German keywords. Without seeing parallel speech-transcriptions or translations, the model achieves a precision at ten of 58\%. We show that most erroneous retrievals contain equivalent or semantically relevant keywords; excluding these would improve $P@10$ to 91\%.
\end{abstract}

\section{Introduction}
\label{sec:intro}

Current automatic speech recognition (ASR) systems are trained on large amounts of transcribed speech audio.
For many languages, it is difficult or impossible to collect such annotated resources~\cite{besacier+etal_speechcom14}.
Furthermore, in contrast to supervised speech systems, human infants acquire language without access to hard labels, and instead rely on other signals, such as visual cues, to ground speech~\cite{roy+pentland_cogsci02,rasanen+rasilo_psych15}.
Recent studies have therefore started to consider how speech models can be trained on 
unlabelled speech paired with images~\cite{synnaeve+etal_nipsworkshop14,harwath+etal_nips16}.
Grounding speech using co-occurring visual context could be a way to build systems when annotations cannot be collected, e.g.\ for endangered or unwritten languages~\cite{chrupala+etal_acl17}.
In robotics, similar methods could be used to learn new words from co-occurring audio and visual signals~\cite{taniguchi+etal_advrob16}.

As in~\cite{harwath+etal_nips16,chrupala+etal_acl17,kamper+etal_interspeech17}, we consider the setting where unannotated images of natural scenes are paired with unlabelled spoken captions.
We specifically build on~\cite{kamper+etal_interspeech17}, which proposed a model that can map speech to text labels: a trained visual tagger is used to obtain soft text labels for each training image, and a neural network is then trained to map speech to these targets.
The result is a model that can be used for keyword spotting, predicting which utterances in a search collection contain a given written keyword.
It does so without observing any parallel speech and text.
In~\cite{kamper+etal_interspeech17}, an English visual tagger was used to ground unlabelled English speech, so English speech was searched using English keywords.

Here we propose an approach where the languages of the speech and visual tagger are not matched, with the aim of performing \textit{cross-lingual keyword spotting}.
Given a textual keyword in one language (the query language), the task is to retrieve speech utterances containing that keyword in another language (the search language). 
{For example}, given the English keyword `doctor', the task could be to search through a spoken Swahili corpus for utterances such as \textit{nataka kuona daktari} (`I need to see a doctor').
While parallel speech-transcriptions and translations are {often} difficult to obtain for {low-resource} languages, a collection of spoken descriptions of images could (potentially) be created by native speakers without writing or translation skills.
We explore whether such paired 
speech-image data is sufficient for training a cross-lingual keyword spotter, thereby bringing together these two strands of research (joint image-speech modelling and cross-lingual retrieval).

Due to the 
lack of suitable resources in truly low-resource languages,  we demonstrate a proof-of-concept implementation where we use German keywords to search through untranscribed English speech. Specifically,
our setup builds on pairs of images and unlabelled English speech, 
and we use a visual tagger producing German text labels as targets for the speech network.
For the task of cross-lingual keyword spotting, a model is given a written German keyword (e.g.\ \textit{Hunde}, the German word for `dogs') and asked to retrieve English speech utterances containing that keyword (e.g.\ `two dogs playing outside near the water').
In extensive analyses, we compare the cross-lingual visual grounding model to several new alternatives (not in~\cite{kamper+etal_interspeech17}).
We find that most errors are due to semantic confusions; adjusting for these brings our model close to a directly supervised system. 

\section{Related work}

Keyword spotting is a well-established task; 
the goal is to retrieve utterances in a search collection that contain spoken instances of a given written keyword~\cite{wilpon+etal_assp90,szoke+etal_interspeech05,garcia+gish_icassp06}.
The query and search languages are the same and typically the aim is to find exact matches. But weaker (semantic) matching has also been studied~\cite{chelba+etal_ieee08,lee+etal_slt12,li+etal_asru13,lee+etal_taslp15}.
In cross-lingual keyword spotting, utterances in one language should be retrieved in response to user text queries in a different language.
Here there has been less research, but early work~\cite{sheridan+etal_sigir97} proposed to cascade ASR with text-based cross-lingual information retrieval~\cite{oard+diekema_arist98}.
This is only possible when transcribed speech are available for building an ASR system. 
Some recent work has proposed models that can translate speech in one language directly to text in another~\cite{duong+etal_naacl16,bansal+etal_eacl17,weiss+etal_interspeech17,berard+etal_icassp18}, but this requires parallel speech with translated text.
We use visual context as supervision for settings where translations are unavailable.

Several recent studies have trained models on images paired with unlabelled speech~\cite{synnaeve+etal_nipsworkshop14,harwath+etal_nips16,chrupala+etal_acl17,drexler+glass_glu17,leidal+etal_asru17,scharenborg+etal_arxiv18,kamper+etal_arxiv17,harwath+etal_icassp2018}.
Most approaches map images and speech into a common space, allowing images to be retrieved using speech and vice versa.
Although useful, labelled (textual) predictions are not possible.
The model of~\cite{kamper+etal_interspeech17} uses an external visual tagger to tag training images with text labels, enabling the model to map speech to text labels (without using transcriptions).
We extend this approach by applying a visual tagger in one language to parallel images and speech in another (the search) language.
To our knowledge, cross-lingual keyword spotting has not been attempted using visual speech grounding.
Finally, recent work has used vision as an additional input modality for (textual) machine translation~\cite{specia+etal_wmt16,elliott+akos_arxiv17,elliott+etal_wmt17}.
We consider \textit{speech} retrieval, with vision as the \textit{only} \mbox{supervisory signal}. 

\section{Model}
\label{sec:model}

\begin{figure}[!b]
    \vspace*{-2pt}
    \centering
    \includegraphics[width=0.95\linewidth]{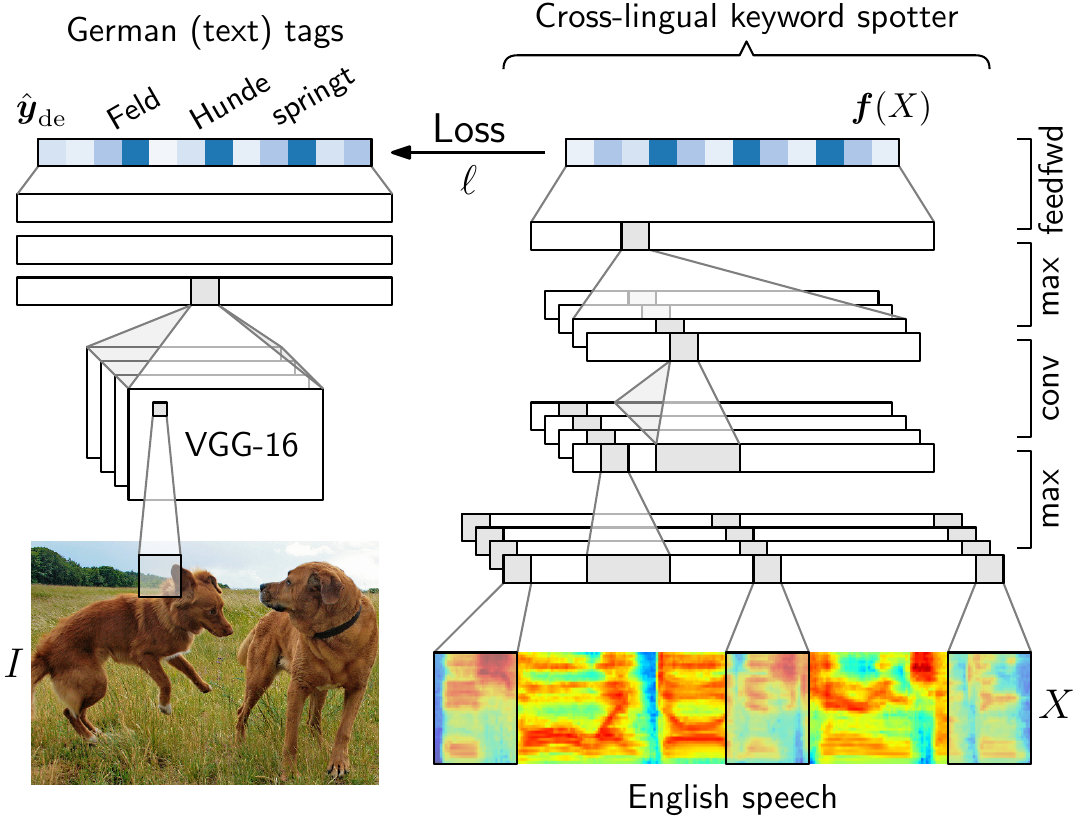}
    \imagesep
    \caption{During training, an external German visual tagger produces text targets for a speech network taking in English speech.
    The speech network is therefore trained using only parallel images and unlabelled English spoken captions.
    Here the visual target $\hat{\vec{y}}_{\textrm{\textup{de}}}$ should ideally be close to $1$ for words such as
\textit{Feld} (\textup{field}), \textit{Hunde} (\textup{dogs}), \textit{springt} (\textup{jump}), and \textit{gr\"{u}n} (\textup{green}).
    }
    \label{fig:xvisionspeech_cnn}
\end{figure}

\begin{figure}[!b]
    \centering
    \includegraphics[width=0.95\linewidth]{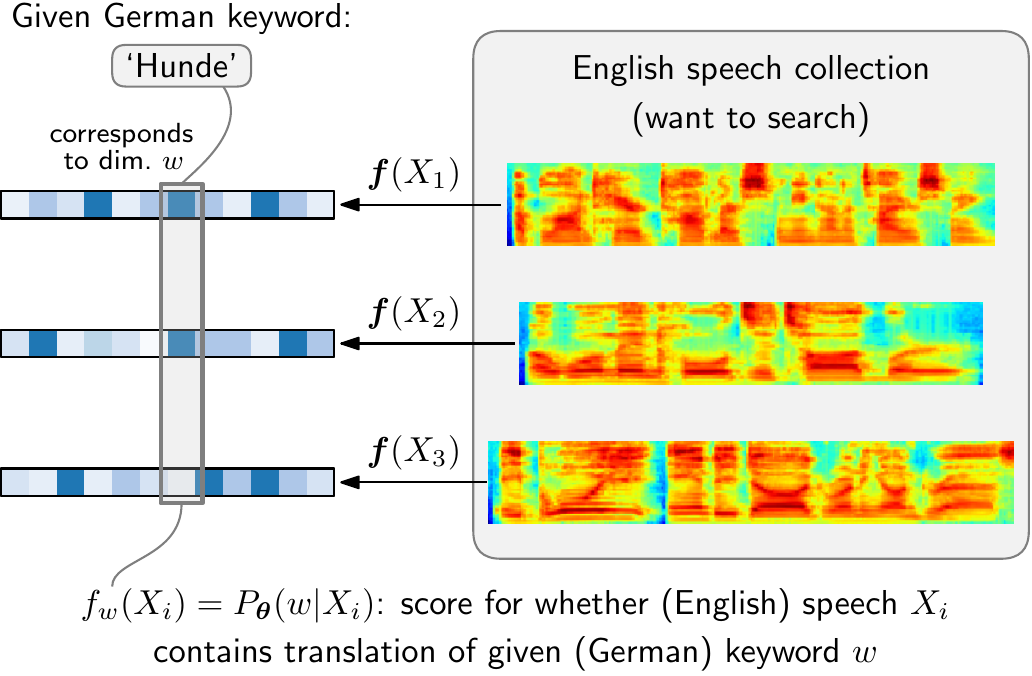}
    \imagesep
    \caption{After training, $\vec{f}(X)$ can be applied as a cross-lingual keyword spotter. An unseen English test utterance is passed through the model, and the resulting output is interpreted as a score for a particular German keyword occurring in the speech.
    }
    \label{fig:xvisionspeech_cnn_test}
\end{figure}

Given an unlabelled corpus of parallel images and spoken captions in the search language (English), we use an external visual tagger in the query language (German) to produce soft targets for a speech network.
This is illustrated in Figure~\ref{fig:xvisionspeech_cnn}, where training image $I$  is paired with English caption $X = \vec{x}_1, \vec{x}_2, \ldots, \vec{x}_T$, with each frame $\vec{x}_t$ an acoustic feature vector, e.g.\ Mel-frequency cepstral coefficients (MFCCs).
Image $I$ is tagged with German text labels, which serves as targets for the speech network $\vec{f}(X)$.
The result is a network that maps English speech to German keyword labels (ignoring order, quantity and where the translations of the keywords occur).
During testing, the model is applied as a cross-lingual keyword spotter as shown in Figure~\ref{fig:xvisionspeech_cnn_test}; each speech utterance in an unseen English search collection is passed through $\vec{f}(X)$, and the output is used to predict whether a given German keyword (text query) is present.
In testing, only English speech input is used (no images). We now give full details.

\subsection{Detailed model description}
\label{sec:model_details}

For training (Figure~\ref{fig:xvisionspeech_cnn}), if we knew the German words occurring in English training utterance $X$, we could construct a multi-hot cross-lingual bag-of-word (BoW) vector $\vec{y}_{\textrm{xbow}} \in \{0,1\}^W$, with $W$ the vocabulary size and each dimension $y_{\textrm{xbow}, w}$ a binary indicator for whether $X$ contains a translation of German word $w$. 
However, we do not have transcriptions or translations to obtain such ideal cross-lingual BoW supervision.
Instead, we only have the image $I$ which is paired with $X$.
Rather than binary indicators, we use a multi-label visual tagging system producing soft targets $\hat{\vec{y}}_{\textrm{de}} \in [0,1]^W$, with $\hat{y}_{\textrm{de}, w} = P(w | I)$ the estimated probability of German word $w$ being relevant given image $I$. 
In Figure~\ref{fig:xvisionspeech_cnn}, $\hat{\vec{y}}_{\textrm{de}}$ would ideally be close to $1$ for $w$ corresponding to words such as
\textit{Feld} (field), \textit{Hunde} (dogs), \textit{springt} (jump), and \textit{gr\"{u}n} (green), and close to $0$ for irrelevant dimensions. 
Note that the visual tagger is assumed to be external: {whereas the speech network $\vec{f}(X)$ is trained, the tagger is held constant}. 

Given $\hat{\vec{y}}_{\textrm{de}}$ as target, we train the speech
model $\vec{f}(X)$ (Figure~\ref{fig:xvisionspeech_cnn}, right).
This model (parameters $\vec{\theta}$) consists of a convolutional neural network (CNN) over the speech $X$ with a final sigmoidal layer so that $\vec{f}(X) \in [0,1]^W$.
We interpret each dimension of the output as $f_w(X) = P_{\vec{\theta}}(w | X)$. 
We train $\vec{f}(X)$ using the summed cross-entropy loss, which (for a single training example) is:
\vspace*{-10pt}
\begin{align}
    \ell(\vec{f}(X), \hat{\vec{y}}_\textrm{de}) 
    &= -\sum_{w = 1}^W \left\{ \hat{y}_{\textrm{de}, w} \log f_w(X) \;\; +\right. \nonumber \\ 
    &\qquad\qquad \left. (1 - \hat{y}_{\textrm{de}, w}) \log\left[1 - f_w(X) \right] \right\}
    \label{eq:summed_cross_entropy}
\end{align}
If we had $\hat{y}_{\textrm{de}, w} \in \{0, 1\}$, as in $\vec{y}_\textrm{xbow}$, this
would be the summed log loss of $W$
binary classifiers.
Note that the size-$W$ (German) vocabulary of the system is implicitly given by the visual tagger.


\subsection{The German visual tagger}

A visual tagger is a multi-label computer vision system that predicts an unordered set of words (nouns, adjectives, verbs) that accurately describes aspects of a scene~\cite{barnard+etal_jmlr03,guillaumin+etal_iccv09,chen+etal_icml13}.
Ideally we want an existing vision system in the query language 
(Figure~\ref{fig:xvisionspeech_cnn}, left).
Although it is fair to assume such a system would be available if the query language is high resource (\S\ref{sec:intro}), 
we could not find an off-the-shelf German tagger. 
We therefore train our own German visual tagger on separate data.

We use the {Multi30k} dataset, which contains around 30k images each annotated with five written German captions~\cite{elliott+etal_wvl16}.
Captions are combined into a single BoW target after removing stop words.
As basis for our tagger, we use VGG-16~\cite{simonyan+zisserman_arxiv14}, trained on around 1.3M images~\cite{deng+etal_cvpr09}, but we replace the final classification layer with four 2048-unit ReLU layers followed by a final sigmoidal layer for predicting word occurrence.
VGG-16 was used in a similar way in previous vision-speech models~\cite{harwath+etal_nips16,chrupala+etal_acl17}.
The visual tagger 
is trained on {Multi30k} with the output layer limited to the $W = 1\textrm{k}$ most common German word types in the captions.
Only the additional fully-connected layers are updated, i.e.\ the VGG-16 parameters are not fine-tuned.

Importantly, none of the training images here overlap with the test data used in our experiments (\S\ref{sec:exp}). Thus, the visually grounded model does not get even indirect access to the (written) German translations, 
so we use it as if it is external.

%


\section{Experiments}
\label{sec:exp}

Our goal is to find spoken utterances in a search language that contain written keywords from a query language.
Our model, referred to as \system{XVisionSpeechCNN}, {does} so without using any transcribed or parallel data; instead, it relies
solely on utterances in the search language that are paired with images, which we process by automatically adding visual keywords in the query language (\S\ref{sec:model}, Figure~\ref{fig:xvisionspeech_cnn}). At test time, our proposed model is given a keyword in the query language and has to retrieve corresponding utterances in the search language, without access to parallel data, transcriptions, or utterance-image pairs (Figure~\ref{fig:xvisionspeech_cnn_test}). 

\subsection{Experimental setup and evaluation}
\label{sec:exp_setup}

We train our visually grounded cross-lingual keyword spotting model \system{XVisionSpeechCNN} 
on the 
Flickr8k Audio Captions Corpus of parallel images and spoken captions containing 8k~images, each with five spoken English captions~\cite{harwath+glass_asru15}.
The audio comprises around 37 hours of non-silent speech, and comes with train, development and test splits of 30k, 5k and 5k utterances, respectively.
We parameterise speech as 13 MFCCs with first and second order derivatives, giving 39-dimensional input vectors.
Utterances longer than 8~s are truncated (99.5\% are shorter).
\system{XVisionSpeechCNN} has the same structure as the monolingual model from~\cite{kamper+etal_arxiv17}:
1-D ReLU convolution with 64 filters over 9 frames; max pooling over 3 units; 1-D ReLU convolution with 256 filters over 10 units; max pooling over 3 units; 1-D ReLU convolution with 1024 filters over 11 units; max pooling over all units; 3k-unit fully-connected ReLU; and the 1k-unit sigmoid output.
We train using Adam~\cite{kingma+ba_iclr15} for 25 epochs with early stopping, a learning rate of $1\cdot10^{-4}$ and a batch size of eight.

For evaluation, we need reference German translations for the English test utterances.
The data set of~\cite{elliott+etal_wvl16} contains German translations for a subset of the English development and test utterance of the Flickr8k corpus. 
We perform evaluation on these utterances, with approximately 1k English utterances for development and 1k utterances for testing, each with one German reference translation.
As keywords, we randomly selected 39 words from the data on which the visual tagger was trained.

When applying \system{XVisionSpeechCNN} to test data (Figure~\ref{fig:xvisionspeech_cnn_test}), we interpret the output $f_w(X) \in [0, 1]$ as a score for how relevant an English utterance $X$ is given German keyword $w$.
The models we compare to (below) also gives this type of scoring.
To obtain a hard prediction from a model, we set a threshold $\alpha$ and label all keywords for which $f_w(X) > \alpha$ as relevant.
By comparing this to the reference translation, precision and recall can be calculated.
We stem the words in both the prediction and reference translation, so that inflections are not marked as errors.
To measure performance independent of $\alpha$, we report \textit{average precision (AP)}, the area under the precision-recall curve as $\alpha$ is varied.
We also consider how a model ranks utterances in the test data from most to least relevant for each keyword~\cite{hazen+etal_asru09,zhang+glass_asru09}: \textit{precision at ten ($P@10$)} is the average precision of the ten highest-scoring proposals; \textit{precision at $N$ ($P@N$)} is the average precision of the top $N$ proposals, with $N$ the number of true occurrences of translations of the word; and \textit{equal error rate (EER)} is the 
rate at which false acceptance and rejection rates are equal.

\subsection{Baselines and comparison models}
\noindent \textbf{\system{DETextPrior}.} This baseline  completely ignores the search language utterance and relies only on unigram probabilities of the keywords in the query language. Comparisons to \system{DETextPrior} indicate how much better our model does than simply predicting common German words for any English speech input.

\vspace*{2pt}
\noindent \textbf{\system{DEVisionCNN}.} One question is whether \system{XVisionSpeechCNN} learns to ignore aspects of the acoustics that are not indicative of visual targets. This baseline
is an attempt to test this: as the representation for each test utterance, it passes through the German visual tagger the \textit{true} image paired with that utterance. If \system{XVisionSpeechCNN} had access to ideal visual tags, then \system{DEVisionCNN} would be an upper bound, but in reality our model could do better or worse (since training does not generalise perfectly).

\vspace*{2pt}
\noindent \textbf{\system{XBoWCNN}.} To check how reliable automatically predicted tags are in comparison to ground truth text labels, we train as an upper bound \system{XBoWCNN}, which has access to the keywords that indeed appear as translations in the search language utterances. 
\subsection{Results}


\begin{table}[!b]
    \mytable
    \caption{Transcriptions of the top English utterances retrieved using \system{XVisionSpeechCNN} for a selection of German keywords. Incorrect retrievals (i.e.\ where the reference translation of the utterance does not contain the keyword) are marked with $*$. Different error types (Table~\ref{tbl:analysis}) are marked with (1),~(2)~and~(3).}
    \tablesep
    \begingroup
    \eightpt
    \begin{tabularx}{\linewidth}{L}
        \toprule
        \textit{Fahrrad} (bicycle) \\
        man riding a bicycle on a foggy day \\
        a biker does a trick on a ramp $*$ \hfill (2) \\
        a person is doing tricks on a bicycle in a city \\
        \midrule
        \textit{Feld} (field) \\
        a team of baseball players in blue uniforms walking together on field \\
        a brown and black dog running through a grassy field $*$ \hfill (1) \\
        two small children walk away in a field \\
        \midrule
        \textit{gro{\ss}(en)} (big) \\
        a large crowd of people ice skating outdoors \\
        a surfer catching a large wave in the ocean \\
        a small group of people sitting together outside~$*$ \hfill (3)\\
        \midrule
        \textit{gr\"{u}n(en)} (green) \\
        boy wearing a green and white soccer uniform running through the grass \\
        a girl is screaming as she comes off the water slide $*$ \hfill (3) \\
        a brown dog is chasing a red frisbee across a grassy field~$*$ \hfill (2) \\
        \midrule
        \textit{Hemd} (shirt) \\
        a woman in a red shirt and a man in white stand in front of a mirror \\
        a man in a blue shirt lifts up his tennis racket and smiles \\
        a man in blue cap and jacket looks frustrated $*$ \hfill (2) \\
        \midrule
        \textit{klettern/klettert} (climbing) \\
        a lone rock climber in a harness climbing a huge rock wall \\
        a man is rock climbing at sunset $*$ \hfill (1)\\
        a man is laying under a large rock in the forest $*$ \hfill (2) \\
        \midrule
        \textit{Personen} (people) \\
        two people are riding a ski lift with mountains behind them \\
        two women are climbing over rocks near to the ocean~$*$ \hfill (2) \\
        two people sit on a bench leaned against a building with writing on it \\
        \midrule
        \textit{Stra{\ss}e} (street) \\
        a woman in black and red listens to an ipod walks down the street \\
        people on the city street walk past a puppet theater \\
        an asian woman rides a bicycle in front of two cars $*$ \hfill (2)\\
        \bottomrule
    \end{tabularx}
    \endgroup
    \label{tbl:eg}
\end{table}

\begin{table}[t]
    \mytable
    \caption{Cross-lingual keyword spotting results (\%) on test data.}
    \tablesep
    \begingroup
    \eightpt
    \begin{tabularx}{\linewidth}{@{\ }Lcccc}
        \toprule
        Model & $P@10$ & $P@N$ & EER & AP \\
        \midrule
        \system{DETextPrior} & \mbox{\hphantom{0}7.2} & \mbox{\hphantom{0}6.3} & 50\hphantom{.0} & 10.4 \\
        \system{DEVisionCNN} & 41.5 & 32.9 & 25.9 & 29.7 \\
        \addlinespace
        \system{XVisionSpeechCNN} & 58.2 & 40.4 & 23.5 & 40.0 \\
        \addlinespace
        \system{XBoWCNN} & 80.8 & 54.3 & 19.1 & 54.3 \\
        \bottomrule
    \end{tabularx}
    \endgroup
    \label{tbl:keyword_spotting}
\end{table}

To first illustrate the cross-lingual keyword spotting task, Table~\ref{tbl:eg} shows example output from \system{XVisionSpeechCNN} for a selection of German keywords with the top English utterances that were retrieved in each case. Utterances where the reference German translation did not contain the given keyword are marked with $*$.
Of the 24 shown retrievals, ten are incorrect.

Table~\ref{tbl:keyword_spotting} shows the results on the test data for \system{XVisionSpeechCNN} and the upper- and lower-bound models.
Without seeing any speech transcriptions or translated text, \system{XVisionSpeechCNN} achieves a $P@10$ of 58\%, with \system{XBoWCNN} the only model to outperform the visually grounded model. 
By comparing performance to \system{DETextPrior}, we see that \system{XVisionSpeechCNN} is not just predicting common German words.
Interestingly, \system{XVisionSpeechCNN} also outperforms \system{DEVisionCNN} over all metrics. If the former were perfectly predicting the German visual tags (which is what it is trained to do), then the performance of these two models would be the same.
We see, however, that \system{XVisionSpeechCNN} is doing more than simply mapping the acoustics to the visual tags; we speculate that it is therefore picking up  information in the speech which cannot be obtained from the corresponding test images.

\subsection{Further analysis}

\begin{table}[t]
    \mytable
    \caption{Analysis of errors by a human annotator of the top ten retrievals on development data.
    Percentages (\%) indicate the absolute drop in $P@10$ due to that error type.}
    \tablesep
    \begingroup
    \eightpt
    \begin{tabularx}{\linewidth}{@{\ }lCCCC}
        \toprule
        & \multicolumn{2}{c}{\system{XVisionSpeech}} & \multicolumn{2}{c}{\system{XBoWCNN}} \\
        \cmidrule{2-3} \cmidrule(l){4-5}
        Error type & Count & \%  & Count & \% \\
        \midrule
        (1) Correct (exact) & \mbox{\hphantom{0}32} & \mbox{\hphantom{0}8.2} & 45 & 11.5 \\
        (2) Semantically related & \mbox{\hphantom{0}86} & 22.1 & 13 & \mbox{\hphantom{0}3.3} \\
        (3) Incorrect retrieval & \mbox{\hphantom{0}35} & \mbox{\hphantom{0}9.0} & 19 & \mbox{\hphantom{0}4.9} \\
        \addlinespace
        Total & 153 & 39.3 & 77 & 19.7 \\
        \bottomrule
    \end{tabularx}
    \endgroup
    \label{tbl:analysis}
    \vspace*{-7.5pt}
\end{table}

\noindent \textbf{Error analysis.} Around 40\% of the utterances in the top ten retrievals of \system{XVisionSpeechCNN} still do not contain the given German keyword in the reference translation. To understand the nature of these mistakes, we asked a German native speaker to annotate 
each error in the top ten retrievals with one of the following categories: (1) the reference does not contain the keyword literally, but an equivalent translation; (2) the utterance does not contain a translation of the keyword, but the retrieval is related in meaning; or (3)~the retrieval is completely incorrect.
Examples of the three types of errors are marked on the right in Table~\ref{tbl:eg}.
Errors of type (1) are normally due to a synonym being used; e.g.\ in Table~\ref{tbl:eg} the erroneous utterance shown for \textit{Feld} (field) is a plausible retrieval as the reference contains the word \textit{Wiese} (meadow). An example error of type (2) can be seen for the keyword \textit{Hemd} (shirt): here, the retrieved utterance does not contain the keyword, but mentions other clothing (cap, jacket).

Errors from both \system{XVisionSpeechCNN} and \system{XBoWCNN} were presented to the annotator in shuffled order.
Table~\ref{tbl:analysis} indicates the absolute penalty in $P@10$ for each error type on development data.
For both models, around 10\% of the retrievals marked as errors are actually correct. 
The bulk of errors from \system{XVisionSpeechCNN} is due to semantically related retrievals. 
These retrievals are marked as errors, but could actually be useful depending on the type of retrieval application.
This is in line with~\cite{kamper+etal_arxiv17}, which showed that visual supervision is beneficial for retrieving non-exact but still relevant utterances in the monolingual case.
If type (1) and type (2) errors are not counted as incorrect, \system{XVisionSpeechCNN} and \system{XBoWCNN} would achieve a $P@10$ of 91\% and 95\%, respectively (but, again, this will depend on the use-case).
We leave a larger analysis, which will also measure recall (not only top retrievals), for future work.
\begin{table}[!t]
    \mytable
    \caption{Cross-lingual keyword spotting results (\%) for different variants of \system{XVisionSpeechCNN} on development data.}
    \tablesep
    \begingroup
    \eightpt
    \begin{tabularx}{\linewidth}{@{\ }Lcccc}
        \toprule
        Model & $P@10$ & $P@N$ & EER & AP \\
        \midrule
        \system{XVisionSpeechCNN} & 60.8 & 39.3 & 23.1 & 38.0\\
        \system{KeyXVisionSpeechCNN} & 60.0 & 39.6 & 24.5 & 36.9 \\ 
        \system{OracleXVisionSpeechCNN} & 57.4 & 37.6 & 24.8 & 36.5 \\ 
        \bottomrule
    \end{tabularx}
    \endgroup
    \label{tbl:variants}
    \vspace*{-7.5pt}
\end{table}

\vspace*{\itemsep}
\noindent \textbf{Variants and ideal supervision.} We compare different variants of \system{XVisionSpeechCNN} to gain insight into properties of the model.
\system{XVisionSpeechCNN} produces scores $\vec{f}(X) \in [0,1]^W$ for all $W = 1\textrm{k}$ words in its output vocabulary. 
But we are actually only interested in those dimensions $w$ corresponding to the test keywords.
If we knew the keywords at training time, we could train a model which only tries to predict the visual tags corresponding to these keywords.
Table~\ref{tbl:variants} shows development performance for such a model, \system{KeyXVisionSpeechCNN}.
Performance is similar to that of \system{XVisionSpeechCNN}, with the latter being slightly better on most metrics.
To understand this improvement, note that \system{XVisionSpeechCNN} can be seen as a variant of \system{KeyXVisionSpeechCNN} trained in a multitask fashion: it is trying to predict extra words not used during testing~\cite{caruana_ml97}.
This effectively regularises our model (improving results).

\system{XVisionSpeechCNN} is trained on soft scores from a visual tagger.
What if we had 
the true hard assignments from the manual annotations for the 
training images?
\system{OracleXVisionSpeechCNN} is trained on such oracle targets.
Table~\ref{tbl:variants} shows that this is actually detrimental.
In~\cite{aytar+etal_nips16}, where video was paired with general audio (not speech), soft targets were also used (as in \system{XVisionSpeechCNN}).
They described this as a student-teacher approach, where the student (in our case the speech network) is trying to distil knowledge from the teacher network (in our case the visual tagger).
It has been shown~\cite{gupta+etal_cvpr16,hinton+etal_arxiv15} that training using soft targets can be beneficial for the student network, which aligns with our findings here. 

\section{Conclusion}

We proposed the first visually grounded speech model for cross-lingual keyword spotting.
By labelling images with tags from a multi-label vision system in the query language (German), we 
train a network that maps unlabelled speech in the search language (English) to German keyword labels. 
Using this network for spotting whether translations of German keywords occur in English speech, we achieve a $P@10$ of almost 60\%.
The majority of errors are due to semantically related retrievals; when these are taken into account, our approach comes close to a supervised model trained on parallel speech with text translations.
In further analysis, we showed that by implicitly predicting tags not in the keyword set, we are getting a small benefit from multitask learning.
We also showed that using soft targets from the visual tagger is better than oracle hard targets; this aligns with findings in student-teacher knowledge distillation studies.
Future work will consider error analyses at a larger scale and applications on truly low-resource (e.g.\ unwritten) languages.

\bibliography{xlingual_flickraudio}

\end{document}